# Audio Frequency-Time Dual Domain Evaluation on Depression Diagnosis

Yu Luo[1], Nan Huang[1], Sophie Yu[2], Hendry Xu[3], Jerry Wang[4], Colin Wang[5],

Zhichao Liu[1]*, Chen Zeng[6]

[1]Chongqing Engineering Research Center of Medical Electronics and Information Technology,

Chongqing University of Posts and Telecommunications, Chongqing, China.

[2]Newport High School, Settle, WA 98006, United States

[3]Millburn High School, Millburn, NJ 07041, United States

[4]Basis Independent McLean, McLean, VA 22102, United States

[5]Cupertino High School, Cupertino, CA 95014, United States

[6]Department of Physics, The George Washington University, Washington, DC 20052, United States

**Abstract:** Depression, as a typical mental disorder, has become a prevalent issue significantly impacting public health. However, the prevention and treatment of depression still face multiple challenges, including complex diagnostic procedures, ambiguous criteria, and low consultation rates, which severely hinder timely assessment and intervention. To address these issues, this study adopts voice as a physiological signal and leverages its frequency-time dual domain multimodal characteristics along with deep learning models to develop an intelligent assessment and diagnostic algorithm for depression based on vocal signals. Experimental results demonstrate that the proposed method achieves excellent performance in the classification task for depression diagnosis, offering new ideas and approaches for the assessment, screening, and diagnosis of depression.

## 1 Introduction

Depression is a mental disorder characterized by a prominent and persistent low mood. Some patients may engage in self-harm or suicidal behaviors, and may also present with symptoms such as delusions and hallucinations [1]-[2]. According to WHO, more than 280 million people worldwide suffer from depression, and the number is increasing year by year [3]. A 2021 study in *The Lancet* indicates that due to the impact of the COVID-19 pandemic, the global prevalence of major depressive disorder and anxiety disorder increased by 28% and 26% respectively in 2020 [4]. WHO predicts that by 2030, depression will become the leading cause of global disease burden. Depression not only affects the physical and mental health of patients, but also leads to high disability and fatality rates, being the largest contributor to disability globally. In China, approximately 280,000 people commit suicide each year, and 40% of them suffer from depression [5]. Currently, issues remain such as low awareness of depression, insufficient investment in its management, and prejudice and discrimination, which prevent patients from receiving adequate attention and treatment. A 2021 study in *The Lancet Psychiatry* shows that most depression patients in China have impaired social functions and an extremely low treatment rate. Among patients diagnosed with depression within one year, only 9.5% have received treatment from health service institutions, 3.6% sought treatment from professional mental health doctors, and only 0.5% received adequate treatment [6].

Traditional diagnostic methods for depression include clinical diagnosis and patient self-assessment. Clinical diagnosis involves doctors having face-to-face interviews with patients and comparing the assessment results with the depressive symptoms described in the *Diagnostic and Statistical Manual of Mental Disorders* [7]. However, there are two major problems. First, there is a significant shortage of professional psychologists in China. By the end of 2020, there were only 50,124 practicing (assistant) psychiatrists (3.55 per 100,000 people), which is far fewer compared to the approximately 95 million depression patients [8]. Second, the diagnostic process is complex, and the results are affected by multiple factors such as the patient's descriptive ability, truthfulness, as well as the doctor's theoretical knowledge and treatment experience, resulting in subjective errors. Patient self-assessment is carried out with the help of depression scales such as PHQ-9, BDI, and HAMD, for preliminary evaluation from dimensions like mood, sleep, and interest [9]-[11]. Nevertheless, this method has its drawbacks. On one hand, due to their own emotions and the level of understanding of depression, patients may have subjective cognitive biases. On the other hand, depression scales are designed based on common symptoms, making it difficult to account for individual differences, and the problem statements are somewhat ambiguous.

To address the limitations of traditional diagnostic methods for depression, researchers have developed more objective indicators such as electroencephalogram (EEG), facial expressions, and voice. EEG is a common method for assessing brain function and is used in the diagnosis of mental illnesses. Orgo et al. extracted graph-theoretic features such as functional connectivity, and applied support vector machines and genetic algorithms to classify 64 subjects, achieving an accuracy of 88.10% [12]. Depression can alter non-verbal behavior. Facial expressions contain a large amount of non-verbal information and are high-information-content characteristic indicators for diagnosing depression. Since patients are less sensitive to emotional stimuli, they show fewer facial expression changes. Li et al. provided emotional stimuli to experimental participants through various tasks, extracted 5 types of facial features, established classification models for different genders, and used multiple classifiers for classification. In the video-watching task, the classification accuracy reached 86.8% for females and 79.4% for males [13].

Speech signals have unique advantages in depression detection. They are easily accessible and non-invasive, similar to the concept of auscultation in traditional Chinese medicine. Research shows that audio features can reflect mental states. Through in-depth analysis and research of a large number of speech samples, researchers have found that the voices of depression patients are usually slow and monotonic. Alghowinem et al. extracted various acoustic features and conducted experiments with different classifiers, finding that the hybrid classifier of GMM and SVM had the best performance [14]. Chen et al. used openSMILE to extract the eGeMAPS feature set, established a decision-tree screening model for depression, and achieved an identification accuracy rate of 83.4% and an F1-score of 80.4% on the MODMA dataset [15]. Rejaibi et al. proposed a deep framework based on recurrent neural networks. The proposed method can effectively detect depression and predict its severity from speech [16]. Xu Zhang et al. introduced transfer learning to enhance the model's ability to distinguish depression-related information by fine-tuning wav2vec [17].

In the field of audio-based depression diagnosis, there are currently a series of crucial issues that need to be resolved. First, the limited amount of labeled data is a prominent bottleneck. It is extremely difficult to obtain large-scale and accurately labeled speech data for depression diagnosis. This leads to a shortage of sufficient effective samples during model training. As a result, the model has difficulty learning the comprehensive and precise relationships between audio features and the depressive state, severely restricting the generalization ability and accuracy of the diagnostic model. Second, irrelevant background factors significantly interfere with the diagnostic results. Factors such as identity information, gender, age, and the speech content itself have no direct connection with the core features of audio-based depression. However, during the actual diagnostic process, they may mislead the model's judgment. For example, differences in speech timbre among genders, speech expression habits of different age groups, and the specific semantic content carried by the speech may all interfere with the model's extraction of audio features that truly reflect the depressive state, thus reducing the reliability of the diagnosis. Third, traditional models such as Convolutional Neural Network (CNN)/Long Short-Term Memory Network (LSTM) are obviously ill-suited for audio-based depression diagnosis. When dealing with audio data, traditional models struggle to effectively capture the complex time-series features and long-term dependencies in audio signals, and are unable to accurately extract the subtle audio features related to depression. At the same time, they have poor robustness against issues such as noise and variability in audio data, and are easily affected by environmental factors, which in turn leads to inaccurate diagnostic results.

In this paper, we conduct a comprehensive exploration of frequency-time dual domain features of audio on depression diagnosis and further . Based on that, a specially designed model, 1D-DCNN is proposed and finetuned for efficient depression diagnosis based on audio signal. The paper analyzes how these features can provide more insights into audio characteristics related to depression in both the frequency and time domains. Furthermore, this paper carries out a detailed evaluation and comparison of audio features in the frequency and time domains. By comparing different time-frequency dual-domain features, it is possible to determine which combinations of features are most effective in differentiating between normal and depressive speech patterns. This time-frequency dual-domain evaluation will contribute to the development of more robust and accurate diagnostic models for audio-based depression detection.

## 2 Methods

### 2.1 Data preprocessing and feature extraction

#### 2.1.1 Data collection

In this study, two datasets were adopted for the evaluation of the proposed methods: Chinese dataset Multi-modal Open Dataset for Mental-disorder Analysis (MODMA), and English dataset Distress Analysis Interview Corpus-Wizard of Oz (DAIC-WOZ) [18]-[20].

The MODMA dataset contains EEG and audio recordings of clinically depressed patients along with carefully matched normal controls. These participants were meticulously diagnosed and selected by professional psychiatrists at the hospital. In this study, only the audio data was utilized. The audio data involves a total of 52 participants. Among them, 23 are diagnosed with depression (16 males and 7 females, aged 18-52 years), and 29 are healthy participants (20 males and 9 females, aged 19-52 years). For each participant, 29 recordings were made and sequentially named from 1 to 29. In total, this resulted in 1508 audio recordings.

The DAIC-WOZ dataset is part of a larger clinical interview corpus called the Distress Analysis Interview Corpus (DAIC). Each audio file of the dataset contains dialogue data between a participant and a virtual agent with a sampling rate of 16 kHz. The dataset contains 189 participants, of which 133 are non-depressed and the rest 56 are depressed. Each participant has one recording with a duration of up to 33 minutes.

#### 2.1.2 Data preprocessing

Before extracting speech features, the data needs to be preprocessed, which includes removing silence segment, pre-emphasis, audio segmentation, framing and windowing.

1) **Removing silence.** The Recording Time (RT) of speech comprises Phonation Time (PT) and Speech Pause Time (SPT). The audio dataset contains silent segments that can impede the subsequent learning process of the dilated convolutional neural network. Therefore, prior to extracting speech features, it is essential to eliminate these silent parts. In this study, we employ pydub and librosa libraries to remove the silent segments from the dataset. Specifically, the average decibel value of the speech is calculated and utilized as the threshold for identifying silence. The detect_nonsilence function in pydub is then used to extract the active speech segments. Subsequently, librosa.effects.remix is applied to splice these active speech segments together.

2) **Pre-emphasis.** Pre-emphasis is a crucial signal processing technique aimed at enhancing the high-frequency components of a speech signal. During the vocalization process, the lips and vocal cords introduce effects that lead to relatively greater attenuation of high-frequency components compared to low-frequency ones. Pre-emphasis serves to counteract this, ensuring that high-frequency information is not overly diminished. The formula for pre-emphasis is as follows. By applying a first-order difference equation to the speech samples, the amplitude of the high-frequency components is effectively increased, thereby optimizing the frequency characteristics of the speech signal for subsequent processing.

$$y(t) = x(t) - \mu x(t-1) \tag{1}$$

where μ is the pre-emphasis coefficient, μ ranges from 0<μ≤1, and μ is 0.97 in the study.

**3) Segmentation.** When dealing with speech sequence data, speech segmentation is a commonly employed preprocessing technique. This operation offers multiple advantages. On one hand, it sets a uniform size for the input of the model, which significantly enhances computational efficiency, enabling the model to process data more efficiently and smoothly. On the other hand, it can increase the number of training samples, providing a richer data foundation for model training and contributing to the improvement of the model's generalization ability. It is worth noting that compared with the methods that rely on semantic content for segmentation in text, this approach is more concise. It does not require additional pruning or other redundant operations. It can encompass all speech segments of a piece of speech to the greatest extent possible, completely preserving the original information of the speech data, and providing more comprehensive and accurate data support for subsequent analysis and modeling. In this paper, the optimal length of speech segmentation is determined through experimental enumeration.

**4) Framing and Windowing.** To improve the audio quality and analysis performance, framing and windowing are also important preprocessing techniques. The speech signal is characterized by its time-varying nature. Nevertheless, within a brief time span, it can be considered as a stable, time-invariant signal. This short-term segment is defined as a frame. To prevent drastic disparities between adjacent frames, an overlap is typically introduced between two consecutive frames. This overlapping approach ensures a smoother transition of information from one frame to the next, reducing the potential for abrupt changes that could disrupt subsequent processing and analysis of the speech signal.

Once the speech signal is partitioned into frames, a window function is applied to each individual frame. Among the most prevalently utilized window functions are the Hamming window and the Hanning window. These functions play a crucial role in enhancing the quality of the framed speech signals. By selectively emphasizing certain frequency components, they effectively enhance harmonics. Additionally, they smooth the boundaries of each frame, thereby reducing edge effects that could otherwise distort the signal's characteristics during subsequent processing stages. This ultimately contributes to a more accurate and reliable representation of the speech signal for further analysis.

The window function used in this study is Hamming window, and the form of Hamming window is shown below.

$$\omega(n) = 0.54 - 0.46\,cos\,\cos\left(\frac{2\pi n}{N-1}\right) \tag{2}$$

where 0≤n≤N-1, N is the length of the Hamming window.

### 2.1.3 Feature extraction

Considering the physiological and physical characteristics of speech chain, multimodal speech features, i.e., Linear Prediction Coefficient (LPC) and Mel Frequency Cepstral Coefficients (MFCC) are extracted from speech signals to represent the speech generation and perception processes, respectively.

**1) Linear prediction coefficient (LPC).** Sound generation can be considered as a source-filter model, where the human vocal tract can be modeled as a digital filter. Suppose the input of the model is $u(n)$, and the output $x(n)$ of the channel model is a linear combination of the model's current input $u(n)$ and its past outputs $x(n-i)$. Mathematically, it can be expressed as:

$$x(n) = \sum_{i=1}^{p} a_i x(n-i) + Gu(n) \tag{3}$$

where i=1,2,3...p, and p is the prediction order; G is the gain factor, and $a_i$ is the channel filter factor. LPC utilizes the temporal correlation of speech signals to estimate future samples by establishing a linear prediction model for the vocal channel. By minimizing the mean square error between the input speech and the predicted speech, the filter coefficient equivalent to the sound channel is obtained through the linear prediction method. The linear prediction model also predicts n speech samples, and with past speech samples as conditions, the linear prediction model is designed as follows:

$$\hat{x}(n) = \sum_{i=1}^{p} \hat{a}_i x(n-i) \tag{4}$$

where $\hat{a}_i$ is obtained by minimizing the error between $x(n)$ and $\hat{x}(n)$. $\hat{a}_i$ provides an estimate of the human channel filter coefficient and is an element of the LPC [21].

**2) Mel Frequency Cepstral Coefficients (MFCC).** MFCC represent a linear transformation of the logarithmic energy spectrum based on the nonlinear Mel scale of sound frequency. The frequency-band division of the Mel-frequency cepstrum is equidistantly partitioned on the Mel scale. It approximates the human auditory system more closely than the frequency bands with linear intervals used in the normal logarithmic cepstrum. To extract MFCC features, first, the signal needs to be framed and windowed. Second, the Discrete Fourier Transform (DFT) is applied to each frame of the signal to obtain the power spectrum of each frame. Subsequently, the Mel-filter bank is used to process the power spectrum. The formula for calculating the Mel value of any frequency when applying the Mel-filter to the signal is:

$$mel(f) = 2595 \log_{10}\left(1 + \frac{f}{700}\right) \tag{5}$$

where f is the frequency (in Hz). After performing a logarithmic transformation on the power, the Discrete Cosine Transform (DCT) is used to obtain the MFCC features. The calculation formula of DCT is as follows:

$$X(k) = \sum_{n=0}^{N-1} x_n * cos \cos\left(\frac{2\pi jnk}{N}\right), k = 1,2,3,\ldots\ldots, N-1 \qquad (6)$$

where $x_n$ is the discrete signal and N is the length of the signal [21].

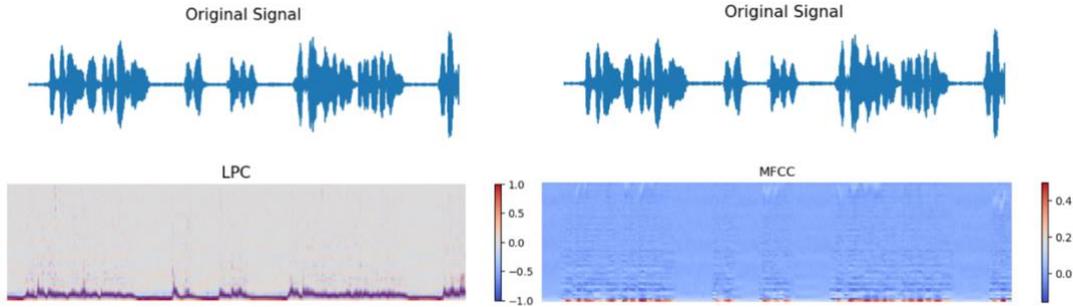

Fig 1. A) LPC; B) MFCC.

**3) Feature size.** In the previous subsection, the temporal length of the features was determined through experiments. Similarly, in order to set a uniform size for the model input, it is necessary to determine the dimension size of the feature extraction in terms of (order/frequency). In this paper, the optimal dimension size of the (order/frequency) features is determined through experiments.

In the study, various durations were tested, ranging from 1 second to 10 seconds, and each speech segment's impact on model training, computational efficiency, and model generalization ability was tested and analyzed. The results are shown in Figure 3-1. During this process, detailed records were kept of the effects of each segment on the model's performance, as well as data on efficiency and generalization. Through repeated testing, careful counting, and in-depth data analysis, the conclusion was reached: when the speech segment duration is 3 seconds, the model achieved relatively ideal performance across all aspects. Therefore, based on experimental counting, the optimal segment length determined in this paper is 3 seconds.

After determining the optimal speech segment length, this paper focuses on exploring the size of the feature dimensions (order/frequency). Since the size of the feature dimension has a crucial impact on the model's performance, rigorous experiments were conducted to find its optimal value. During the experimental process, a counting method was used to observe and record the model's performance across different feature dimensions (order/frequency) in multiple dimensions. The results are shown in Figure 3-2. Through repeated testing, this paper identified the feature dimension (order/frequency) range that allowed the model to achieve optimal performance. Ultimately, it was determined that when the feature dimension (order/frequency) was around 60, the model reached its best state in terms of training stability, computational resource utilization, and data generalization ability. In conclusion, this paper established that the optimal feature dimension size is around 60, providing theoretical basis and parameter support for subsequent model optimization and application.

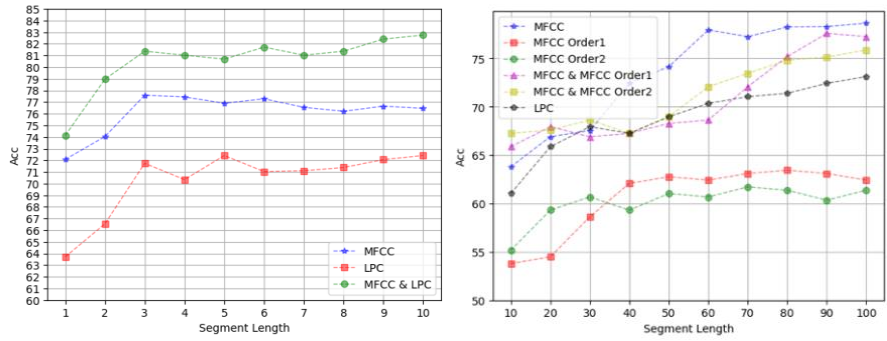

Fig.2 Optimal feature size. A) Time domain (s); B) Frequency domain.

**2.2 1D feature construction and evaluation**

In this study, we specifically focus on the contribution of 1D features within time-frequency dual domain of audio characteristics to the diagnosis of depression and conducts an in-depth and comprehensive exploration. Audio features refer to a set of parameters extracted from raw audio signals, which can characterize the properties of the audio. These features contain rich audio information and play a crucial role in the analysis and understanding of audio. For traditional images, pixels exist in spatial relationships within local neighborhoods in the form of semantic structures. Therefore, two-dimensional convolutional kernels in convolutional neural networks are particularly well-suited for learning local patterns in image data, as pixels are highly correlated within local neighborhoods.

However, when speech data is represented as two-dimensional images (e.g., Mel-frequency cepstral coefficients and linear predictive coefficients), it does not necessarily exhibit similar semantic structures within local neighborhoods. The main reason behind this is that when performing speech feature extraction such as MFCC/LPC, the pixel values on the Y-axis (Frequency/Order, denoted as F) correspond to coefficients for different frequencies/orders, while the pixel values on the X-axis (Time, denoted as T) represent coefficients for different time windows. There is not necessarily a coherence between these two dimensions. In this section of the study, feature processing is carried out along the frequency (F) and time (T) dimensions, using average pooling to obtain a one-dimensional feature vector for subsequent analysis.

When conducting an in-depth exploration of the obtained 1D features, we meticulously selected a series of classic and highly efficient machine learning methods, including Random Forest (RF), LightGBM (LGBM), Support Vector Machine (SVM), and Extreme Gradient Boosting (XGBoost). To fully unleash the potential of these methods, we use GridSearchCV to find the optimal parameter combinations for each model. We pre-set the parameters to be adjusted and their value ranges, conduct a comprehensive search, and evaluate each parameter combination using cross-validation to select the optimal one. Through this rigorous process, we obtain the best configurations of each method for processing 1D audio features, achieving the optimal results of machine learning and accurately revealing the underlying patterns in the audio data.

To further explore the diagnosis potential of 1D audio features, this study also proposes RSENet, a deep neural network using Residual and Squeeze-and-Excitation blocks [22]. With deeper architectures and high-dimensional feature representations, large neural networks are susceptible to the vanishing gradient problem, where the gradient signal diminishes across layers. Residual connections introduce skip pathways that allow gradients to flow directly to earlier layers, bypassing the mandatory backpropagation that would usually diminish the gradient. This design enables the model to learn residual mappings — differences between the input and output representations — rather than complete transformations [23]. We also use Squeeze-and-Excitation blocks to perform dynamic channel-wise feature recalibration. This mechanism involves two operations: the squeeze step, which aggregates global information via average pooling, and the excitation step, which generates a set of adaptive scaling coefficients through fully connected layers with sigmoid activations. These learned coefficients modulate the importance of each feature dimension, simulating a primitive attention mechanism.

In the fields of machine learning and data analysis, feature importance evaluation is a crucial task. It provides strong support for understanding the internal structure of data, optimizing model performance, and making informed decisions. In complex machine learning models, especially black-box models like deep neural networks, it is extremely difficult to intuitively understand how the model makes decisions. By evaluating feature importance, we can identify which input features have a greater impact on the model's output results.

In this study, to deeply unearth the crucial information concealed within audio data and thereby precisely grasp the essential characteristics of audio content, we employed the method of Random Forest importance assessment to conduct a systematic and comprehensive exploration of audio features. By constructing a Random Forest model composed of multiple decision trees and inputting audio features into it, we utilized the model's powerful learning and analytical capabilities to quantitatively evaluate the importance of each audio feature within the entire dataset. This enabled us to identify the core features that play a pivotal role in audio analysis, laying a solid foundation for subsequent research work based on audio features.

**2.3 1D-DCNN for Frequency-Time dual domain audio signal**

Given the significantly higher contribution of the frequency-domain of MFCC/LPC feature to depression diagnosis than time-domain, a model called 1D dilated convolutional neural network (1D-DCNN) is proposed in this study. The basic architecture of 1D-DCNN is shown in the following figure. The convolutional process used in this study is a one-dimensional dilated convolutional kernel. Compared with traditional convolutional layers, the dilated convolutional layers have a larger field of view when the number of parameters is the same. Therefore, the dilated convolutional layers can replace the traditional convolutional layers to expand the receptive field in the network, without causing the data loss problems that are brought about by the traditional convolutional layers.

Hyperparameter Grid Search is a method in machine and deep learning for finding the best hyperparameter combinations. Hyperparameters are set before model training and not learned during it. Both Grid Search and Optimization aim to enhance model performance. By finding the best hyperparameters, we can prevent overfitting and underfitting.

In the process of in-depth research on the optimization and improvement of the network architecture, we plan to conduct a comprehensive and meticulous exploration of the network from multiple key dimensions. Specifically, we will start from the following three highly representative perspectives: First, we will conduct an in-depth analysis and comparison between 1D-DCNN and 2D-CNN+pooling (2D Convolutional Neural Network combined with pooling operation), two different network structure models. We will study their differences and advantages in aspects such as feature extraction, data processing, and model performance. Second, we will focus on examining the impact of setting the kernel size to 3/5/7 respectively on the convolution effect, receptive field range, and overall computational complexity of the network. Third, we will conduct an in-depth exploration of the key parameter Dilation rate. We will analyze how it changes the dilated convolution characteristics of the network under different values, and further, how it exerts varying degrees of influence on the network's performance and generalization ability. Through in-depth exploration of these three dimensions, we expect to provide a more solid theoretical basis and practical experience for the optimization of the network architecture.

**2.4 Evaluation metrics**

**1)** Dataset spilt

In order to ensure the fairness and generalizability of model evaluation in audio dataset, the division of training set and test set is very critical, and the data contained in the training set should not appear in the test set, which prevents the model from directly memorizing the speech features of specific individuals to obtain too high scores. Therefore, when dividing data sets for tasks such as speech disorder diagnosis, speaker recognition, or sentiment analysis, it is important to ensure that the samples between the two parts do not overlap.

In the MODMA dataset of this study, each participant contains 29 audio files, so the experimental partitioning dataset should treat each person as an independent sample, rather than each audio file as an independent sample. In this study, 42 participants were selected in the training set and 10 participants were selected in the test set (5 normal participants, 3 male and 2 female; 5 depressed participants, 3 male and 2 female).

**2) Model performance evaluation**

In this study, accuracy and F1-score are adopted for the evaluation of diagnosis performance. Accuracy is a commonly used model evaluation index, which represents the proportion of correct samples in the total number of samples in the model prediction results. F1-score is an indicator used to measure the accuracy of binary classification (or multi-task binary classification) models in statistics. It takes into account both the

Precision and Recall of the classification model, and the high-er the value, the better the model.

## 3 Results

### 3.1 1D feature & ML models evaluation

In this study, the focus is on the one-dimensional features in audio characteristics, with an in-depth and comprehensive exploration. Audio features refer to a set of parameters extracted from raw audio signals that characterize the properties of the audio. These features contain rich information about the audio and play a crucial role in its analysis and understanding. As a special form, one-dimensional features have unique research value. In visual images, pixels exist in spatial relationships within local neighborhoods in the form of semantic structures. Therefore, the two-dimensional square convolution kernels in convolutional neural networks are particularly well-suited for learning local patterns in image data because pixels are highly correlated within local neighborhoods. However, when speech data is represented as two-dimensional images (such as Mel-frequency cepstral coefficients and linear prediction coefficients), they do not exhibit similar semantic structures in local neighborhoods. The main reason behind this is that when extracting speech features like MFCC/LPC, the pixel values on the Y-axis (Frequency/Order, denoted as F) correspond to coefficients at different frequencies/orders, while the pixel values on the X-axis (Time, denoted as T) represent coefficients at different time windows, and there is no inherent continuity between these two dimensions. In this subsection, the feature processing is carried out separately along the frequency (F) and time (T) dimensions, using average pooling to obtain one-dimensional feature vectors for further analysis.

### 3.1.1 Unsupervised Learning Exploration of Speech One-Dimensional Features

To explore the underlying patterns of the raw features, this section first conducts an unsupervised learning exploration. By applying the t-SNE (t-distributed Stochastic Neighbor Embedding) algorithm for dimensionality reduction, the resulting visualization map is carefully observed and analyzed in-depth. As shown in Figure 3-4, it can be observed that the raw speech features and the one-dimensional feature vectors constructed based on the time (T) dimension are not effective in distinguishing between depression patients and normal individuals (first and third columns). However, the one-dimensional feature vectors constructed based on the frequency (F) dimension demonstrate remarkable effectiveness in distinguishing between depression patients and normal individuals (second column). These feature vectors can clearly and distinctly differentiate these two groups with high recognition, providing strong support for subsequent research and applications.

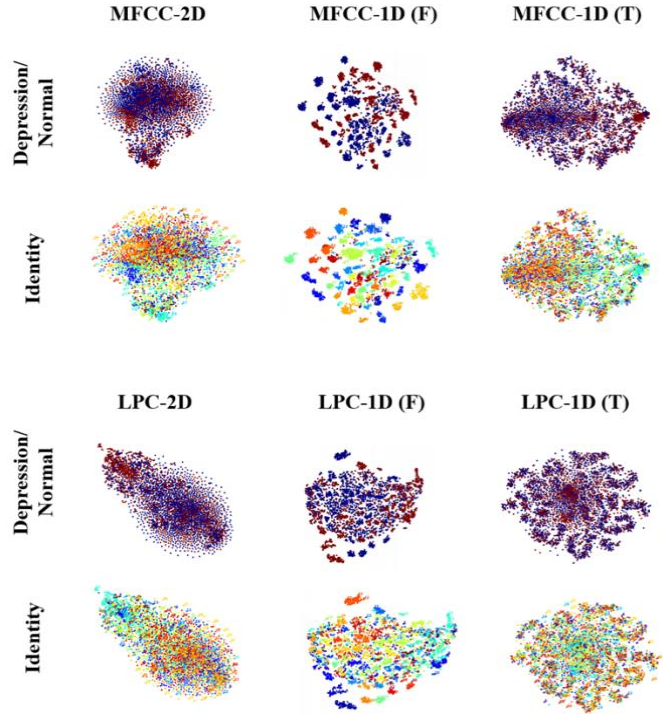

Fig.3 Exploration of speech feature dimensionality reduction

In light of this, this study aims to deeply explore the hidden depressive features in the frequency (F) dimension of LPC and MFCC features, intending to break through the analytical challenges encountered after transforming speech data into two-dimensional images. This will provide strong support for subsequent precise identification and research. The time (T) dimension will also undergo similar analysis and evaluation for comparison. To ensure that the LPC and MFCC features have equal dimensions and to account for the conclusions in Figures 3, an average pooling operation along the time dimension (with a sliding window size of 3 and a stride of 2) is applied to the raw speech features. This results in a 64×64 dimension feature matrix, ensuring computational efficiency and stability for subsequent feature analysis.

### 3.1.2 Machine Learning Evaluation of One-Dimensional Speech Features

To reveal the distribution patterns in the time-frequency feature space of speech, this study constructs an analysis framework based on machine learning, using four machine learning methods: Random Forest (RF), Light Gradient Boosting Machine (LGBM), Support Vector Machine (SVM), and Extreme Gradient Boosting (XGBoost). Table 1 shows the classification results (Accuracy and F1-score) based on one-dimensional speech features, where T represents the time-domain features and F represents the frequency-domain features.

Table 1 Machine learning results of time-frequency and dual-domain speech features

| Feature | RF | LGBM | SVM | XGBoost |
| --- | --- | --- | --- | --- |
| MFCC (F) | 75.31±1.20 | 75.34±0.99 | **76.08±1.25** | 73.60±1.01 |
| | 77.79±0.97 | 77.67±0.93 | **78.14±0.89** | 73.40±1.05 |
| MFCC (T) | 55.13±0.93 | 55.66±1.59 | 55.04±0.64 | 54.54±0.73 |
| | 54.04±1.16 | 54.43±2.20 | 48.98±0.98 | 54.48±0.76 |
| LPC (F) | 72.27±0.72 | 70.72±1.21 | 70.07±2.67 | 71.41±1.01 |
| | 72.74±0.70 | 72.51±1.57 | 71.49±2.10 | 71.32±1.02 |
| LPC (T) | 52.32±0.43 | 52.31±1.06 | 48.67±0.36 | 52.70±0.62 |
| | 57.05±0.85 | 57.32±1.27 | 31.83±1.57 | 51.95±0.62 |

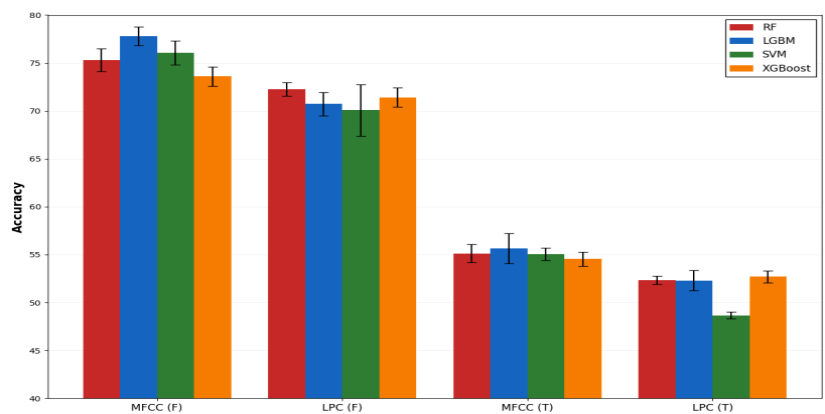

Fig.4 Error bar chart of different speech features in the time-frequency dual domain

From the error bar chart in Figure 4, it can be observed that, whether using MFCC features or LPC features, the feature vectors based on the frequency dimension consistently outperform those based on the time dimension in terms of performance.

This study uses a trained machine learning model to predict each frame of the speech features one by one. The model is trained on a large dataset and is capable of identifying and extracting hidden patterns in the speech features. As a result, it can generate corresponding prediction results for each frame of speech features. Subsequently, the prediction results for each frame are mapped back to the original speech signal, establishing a correspondence between the prediction results and the original speech segments, as shown in Figure 5. To present the analysis results more intuitively, color coding is applied to annotate the original speech. Different colors represent the disease status and non-disease status. This allows the observer to easily and clearly distinguish which segments of the original speech correspond to the diseased state and which correspond to the non-diseased state. This visualization annotation method significantly enhances the readability and comprehensibility of the analysis results, facilitating the quick and accurate interpretation of the health information reflected in the speech data.

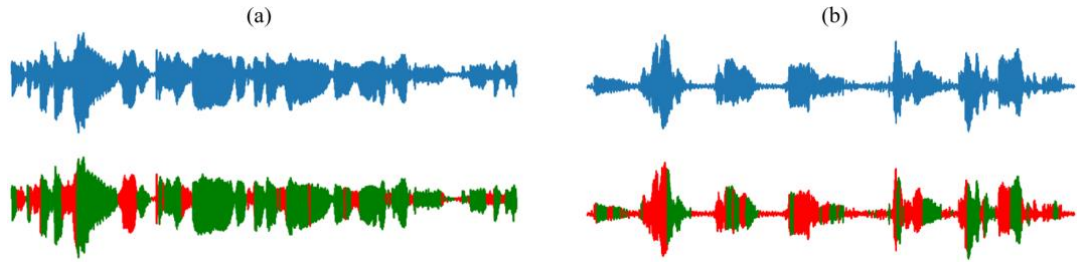

Fig.5 Schematic diagram of the labeling of the depressive/normal state of the original speech. (a) Normal ; (b) Depressed patients.

The trained RSENet model achieved an overall accuracy of 91.28%. The classification report further emphasizes the accuracy of the model. Class 0 (not depressed) achieved a precision of 0.922 and a recall of 0.897, while class 1 (depressed) achieved a precision of 0.905 and a recall of 0.928. The resulting F1-scores, 0.909 for class 0 and 0.916 for class 1, suggest that the network maintains a show balance between precision and recall across both categories.

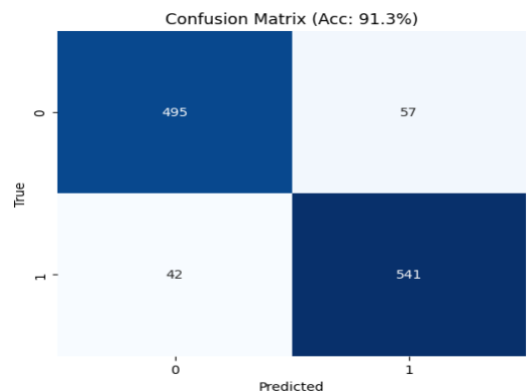

Fig.6 Performance and confusion matrix of RSENet model.

### 3.1.4 Evaluation of the Importance of 1D Frequency Domain Features in Speech

To accurately analyze the significance of each order of one-dimensional frequency domain features in speech, this paper uses the Random Forest importance evaluation method to investigate the one-dimensional frequency domain features of speech, as shown in Figure 3-8. First, the speech dataset is divided into five folds, and then the importance of the one-dimensional frequency domain features is evaluated for each fold of data (Figure 7(a)), followed by Z-Score normalization (Figure 7(b)). From the result graph, it can be concluded that the frequency domain orders with significantly higher feature importance are as follows: 11/12/49/9/22/23/55 (counting from order 0). The related conclusion is consistent with previous studies in this field, where the first 13 MFCC features are often selected.

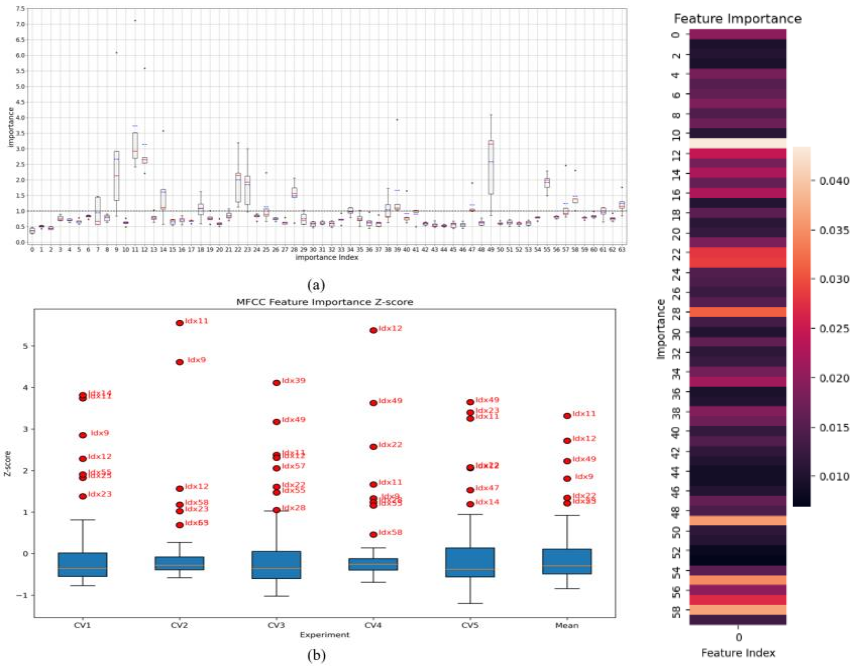

Fig.7 Evaluation of the importance of one-dimensional frequency domain features of speech

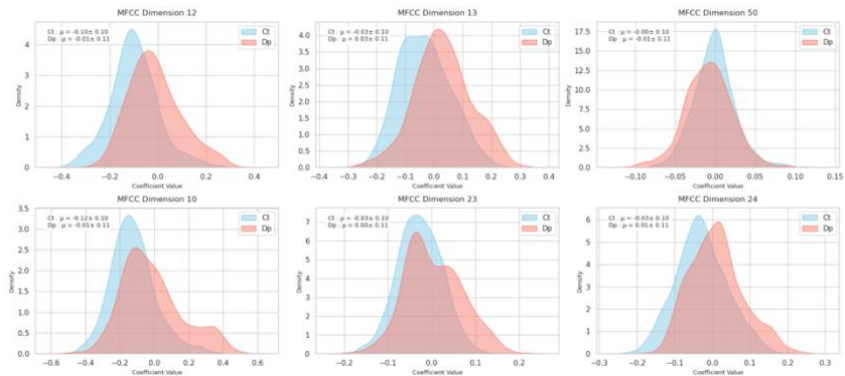

Fig.8 Distribution of the MFCC frequency-domain features with highest importance.

## 3.2 Exploration of 1D-DCNN in the Time-Frequency Dual Domain

The following section will focus on an in-depth analysis and comparison of two different network architecture models: the one-dimensional dilated convolutional neural network (1D-DCNN) and the two-dimensional convolutional neural network with pooling operations (2D CNN + Pooling), in order to find the optimal parameter settings for the networks. In this study, we will systematically investigate the impact of setting the convolution kernel sizes to 3 and 5 on the network's convolutional effects, receptive field size, and computational complexity. Different convolution kernel sizes will result in changes in the granularity and range of feature extraction, as well as affect the network's computational complexity.

Regarding the cases of kernel sizes 3 and 5: a smaller kernel size may capture more detailed information, but is more sensitive to noise. A larger kernel size may lose some fine details but can capture patterns over a larger range. In this paper, it is essential to investigate the impact of these two cases on the receptive field size in both one-dimensional and two-dimensional environments. Additionally, experiments with different numbers of neural network layers will be conducted. In this paper, experiments will be carried out with neural networks having 2, 4, and 6 layers, combined with different network settings: (1) Pure 1D-F: One-dimensional dilated convolution along the frequency direction; (2) Pure 1D-T: One-dimensional dilated convolution along the time direction; (3) 1D-F&2D: One-dimensional dilated convolution along the frequency direction combined with a two-dimensional convolutional neural network with pooling; (4) 1D-T&2D: One-dimensional dilated convolution along the time direction combined with a two-dimensional convolutional neural network with pooling; (5) Pure 2D: Two-dimensional convolutional neural network with pooling. The experimental results are shown in the table below.

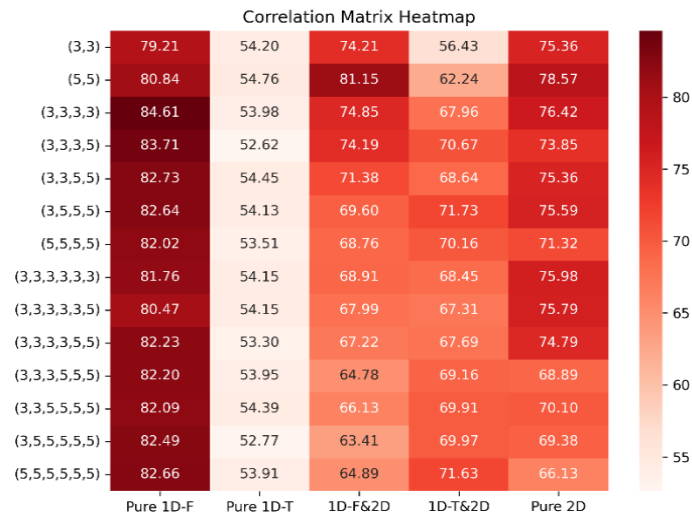

Fig.9 Different parameter settings result in heatmap

From Fig 9, it can be seen that when using fused features, the best performance is achieved with a 4-layer network and a convolution kernel size of 3 along the frequency direction using one-dimensional dilated convolution (i.e., Pure 1D-F). The table shows that the performance of the one-dimensional dilated convolutional neural network along the frequency direction is significantly better than that of the one-dimensional dilated convolutional neural network along the time direction or the standard two-dimensional convolutional neural network. This suggests that disease-related features are likely mainly contained in the frequency domain, while variations in the time domain mainly reflect speech content. Additionally, replacing standard pooling with dilated operations and combining fused features with the one-dimensional dilated convolutional neural network along the frequency axis significantly improved overall performance.

For the network depth and convolution kernel exploration, the dilation rate was set to 2. To further investigate the impact of dilation rate on model performance, the next step will be to conduct relevant exploratory experiments. The network structure used in these experiments is the one-dimensional dilated convolution network along the frequency direction, which consists of 4 layers, with the kernel size set to 3 for each layer. In this exploration, dilation rates of 2 or 3 are set, and the performance of the model under different dilation rates is compared to gain a deeper understanding of the role of the dilation rate parameter in the network's performance. The experimental results are shown in the table below.

Table 2 Experimental results of different expansion rates of the model

| Dilated rate & Layers | Pure 1D-F |
|---|---|
| (2,2,2,2) | 84.61±0.86 |
| | 85.46±0.61 |
| (2,2,2,3) | **86.01±0.52** |
| | **86.87±0.60** |
| (2,2,3,3) | 83.07±0.55 |
| | 84.02±0.56 |
| (2,3,3,3) | 82.96±0.49 |
| | 83.11±0.54 |
| (3,3,3,3) | 81.62±0.52 |
| | 82.66±0.43 |

From Table 2, it can be seen that adjusting the dilation rate of the model has a certain impact on the experimental results. Specifically, after conducting experiments with different dilation rate settings, it was found that when the dilation rate for the first three layers was set to 2 and the dilation rate for the last layer was set to 3, the model achieved the best performance, yielding the most optimal results across all evaluation metrics.

**3.3 Feature ablation study**

In order to explore the potential information in audio, this paper employs spectral masking techniques in ablation experiments to further investigate audio features. By combining the application of spectral masking with a multi-dimensional evaluation approach, this study aims to gain a deeper insight into audio features and the ability for the 1D-DCNN model to generalize on suboptimal data. By masking out random time and frequency regions, 1D-DCNN can't overfit to narrow acoustic cues and must learn from a more robust set of features. This effort is intended to provide a solid theoretical foundation and practical experience for future applications in fields such as depression diagnosis and disease diagnosis [24]-[25].

This paper will perform masking operations on the time (T) and frequency (F) dimensions of the features separately, as shown in Figure 10. The specific strategies are as follows: 1. Time dimension masking: Masking speech features only in the time direction; 2. Frequency dimension masking: Masking speech features only in the frequency direction; 3. Time-frequency bi-dimensional masking: Masking speech features simultaneously in both the time and frequency directions.

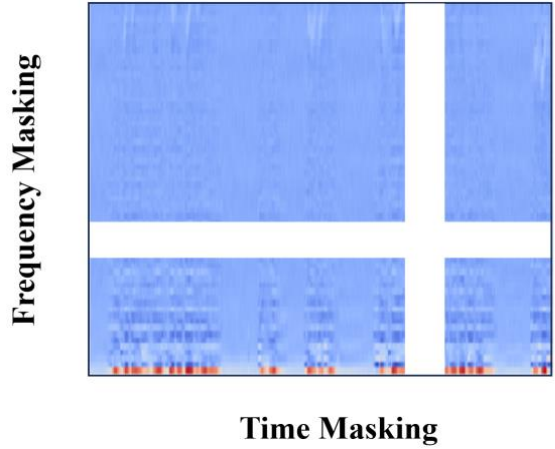

Fig 10 Time-frequency and dual-domain masking

Through these three strategies, this paper will further explore the relationship between speech features in the time and frequency dimensions, revealing the underlying potential information. Table 3 below shows the results obtained from different speech features using the three different strategies.

Table 3 Experimental results of time-frequency dual domain masking with different features

| Feature | Original (Acc/F1-Score) | T Mask (Acc/F1-Score) | F Mask (Acc/F1-Score) | T-F Mask (Acc/F1-Score) |
|---|---|---|---|---|
| LPC | 74.52±0.77 | 73.75±0.58 | 72.56±0.89 | 71.98±0.27 |
|  | 75.28±1.06 | 74.16±0.51 | 72.75±0.42 | 71.95±0.35 |
| MFCC | 82.41±0.52 | 81.16±0.88 | 76.46±0.81 | 74.32±0.53 |
|  | 82.57±0.45 | 81.76±0.53 | 77.42±0.89 | 74.75±0.97 |
| Fusion | **86.01±0.52** | 82.69±0.65 | 81.29±0.86 | 79.56±0.54 |
|  | **86.87±0.60** | 83.20±0.82 | 82.32±1.06 | 80.62±0.99 |

From the analysis of the data in Table 3, it can be observed that both masking speech features in the time direction and in the frequency direction have an impact on recognition accuracy. Specifically, a comparison of the effects of the two masking methods reveals that masking along the frequency direction has a greater impact on recognition accuracy than masking in the time direction. This result indicates that, within speech features, the frequency dimension has a more significant influence on recognition accuracy compared to the time dimension, highlighting the importance of the frequency dimension in speech features.

### 3.4 Results on Other Speech Datasets

In addition, to validate the generalizability of the model, this paper uses the DAIC-WOZ dataset for model evaluation. A grid search for the 1D dilated convolutional network was conducted to obtain the best network parameters. The optimal model parameters obtained were then applied to the DAIC-WOZ dataset for further performance validation and analysis. The data processing and feature extraction process remains consistent with the previous one, and the results are shown in Table 4.

Table 4 Performance on DAIC-WOZ dataset

| Method | Feature | F1-Score |
| --- | --- | --- |
| DepAuidoNet | Spectrogram & Mel-scale filter bank feature | 0.52 |
| FrAUG | Fusion Feature | 0.47 |
| SIDD | MBF | 0.60 |
| DRS | MFCC | 0.62 |
| Proposed | LPC-MFCC | **0.63** |

The results from Table 4 show that, compared to other methods in the literature, the method proposed in this paper achieves the best results on the DAIC-WOZ dataset. A comparison between Tables 2 and 4 reveals that the F1-Score for the DAIC-WOZ dataset is significantly lower than that for the MODMA dataset. One potential explanation is that, the participant score distribution in DAIC-WOZ is concentrated in the non-depressed or mildly depressed group, while the participant score distribution in MODMA is more dispersed. The concentrated distribution in the DAIC-WOZ dataset may have a potential impact on model training. Additionally, the DAIC-WOZ dataset only uses the PHQ-8 scale to binarize the subject labels, which is not sufficiently rigorous in clinical diagnosis, as the threshold for depression should not be determined by a single scale alone. In contrast, the labeling of the MODMA dataset is scientifically rigorous, using multiple scales and doctor diagnoses for labeling. Thirdly, the DAIC-WOZ dataset only covers a single task, namely the interview task. In contrast, the MODMA dataset consists of multiple tasks, including interview, word reading, article reading, and picture description. Multiple tasks provide a more comprehensive assessment of the behavior of depressed patients in various scenarios. Compared to the single interview task in the DAIC-WOZ dataset, the MODMA dataset offers broader scenario adaptability and richer information collection dimensions for assessing depressed patients.

## 4. Discussion and Conclusion

In this study, 1D-DCNN model together with multimodal audio features is proposed for depression diagnosis. LPC and MFCC features are extracted according to the structure of the speech chain to simulate the speech generation and speech perception process. Furthermore, 1D dilated convolutional kernel along the Frequency-axis instead of standard 2D CNN is proposed to extract depression-related information in the frequency domain. The proposed method is evaluated on MODMA and DAIC-WOZ datasets and compared with other current approaches, which proves the effectiveness and accuracy of our proposed model in depression diagnosis based on speech signals.

Nevertheless, there are also some limitations in this study. First of all, only speech signal is used for the diagnosis, while the EEG information is neglected. On one hand, EEG data is harder to obtain than audio in most previous studies and clinical practice, which indirectly results in the situa- tion that most currents approaches use audio data only. On the other hand, EEG has been proved to be an important auxiliary information source for depression diagnosis. Thus, more efficient multimodal feature fusion algorithms are still needed for more accurate depression diagnosis. Secondly, 1D convolution along Frequency axis has been proved to show better performance than standard 2D convolution on the depression diagnosis task based on MFCC- LPC speech features. Our conjecture is that the disease-related features may be mainly contained in the frequency domain rather than the time-domain, which is different from the traditional 2D images where both dimension stand for a spatial domain and contribute equally to the contents. However, the differences of the frequency pattern between audio of normal and depressed people remain to be revealed. Third, in the audio-based depression diagnosis task, whether the diagnostic model indeed learns depression-related subtle features, or only some obvious information such as gender or identity, is still not clear. According to previous studies and our experiments (not shown in this paper), if the subjects are mixed between the training and test datasets of speech segments, the diagnosis accuracy is over 90% [26]. However, if the subjects of the speech in the test dataset are not included in the training set (which is the case for this study), then the best accuracy would dramatically drop to around 80% (~84% for our pro- posed model). Thus, a model that could extract and evaluate on only the disease-related audio features is promised to boost the diagnosis performance. The limitations discussed above could be potential future directions of this area. Overall, effective fusion approaches for multimodal physiological signals, together with more advanced disease-related feature extraction and discriminant models, are still to be further explored. Meanwhile, more lightweight and efficient models that can be easily deployed on mobile devices will promote the practical and clinical use of audio-based depression screening and diagnosis.